\documentclass[10pt,twocolumn,letterpaper]{article}

\usepackage{iccv}
\usepackage{times}
\usepackage{epsfig}
\usepackage{graphicx}
\usepackage{amsmath}
\usepackage{amssymb}
\usepackage{enumitem}
\usepackage{booktabs}
\usepackage{multirow}
\usepackage{verbatim}

\usepackage[pagebackref=true,breaklinks=true,letterpaper=true,colorlinks,bookmarks=false]{hyperref}
\iccvfinalcopy 

\def\iccvPaperID{3911} 

\ificcvfinal\fi

\begin{document}

\title{Joint Inductive and Transductive Learning for Video Object Segmentation}

\author{Yunyao~Mao$^{1}$~~~~~Ning~Wang$^{1}$~~~~~Wengang~Zhou$^{1,2,}$\thanks{Corresponding authors: Wengang Zhou and Houqiang Li}~~~~~Houqiang~Li$^{1,2,}$\footnotemark[1] \\
	{\normalsize $^{1}$ CAS Key Laboratory of Technology in GIPAS, EEIS Department, University of Science and Technology of China} \\
	{\normalsize $^{2}$ Institute of Artificial Intelligence, Hefei Comprehensive National Science Center} \\
	{\tt\small \{myy2016,wn6149\}@mail.ustc.edu.cn, \{zhwg,lihq\}@ustc.edu.cn}
}

\maketitle
\ificcvfinal\thispagestyle{empty}\fi

\begin{abstract}
Semi-supervised video object segmentation is a task of segmenting the target object in a video sequence given only a mask annotation in the first frame. The limited information available makes it an extremely challenging task. Most previous best-performing methods adopt matching-based transductive reasoning or online inductive learning. Nevertheless, they are either less discriminative for similar instances or insufficient in the utilization of spatio-temporal information. In this work, we propose to integrate transductive and inductive learning into a unified framework to exploit the complementarity between them for accurate and robust video object segmentation. The proposed approach consists of two functional branches. The transduction branch adopts a lightweight transformer architecture to aggregate rich spatio-temporal cues while the induction branch performs online inductive learning to obtain discriminative target information.
To bridge these two diverse branches, a two-head label encoder is introduced to learn the suitable target prior for each of them. The generated mask encodings are further forced to be disentangled to better retain their complementarity. 
Extensive experiments on several prevalent benchmarks show that, without the need of synthetic training data, the proposed approach sets a series of new state-of-the-art records. Code is available at \url{https://github.com/maoyunyao/JOINT}.
\end{abstract}

\section{Introduction}

\begin{figure}[t]
	\begin{center}
	\includegraphics[width=1.0\linewidth]{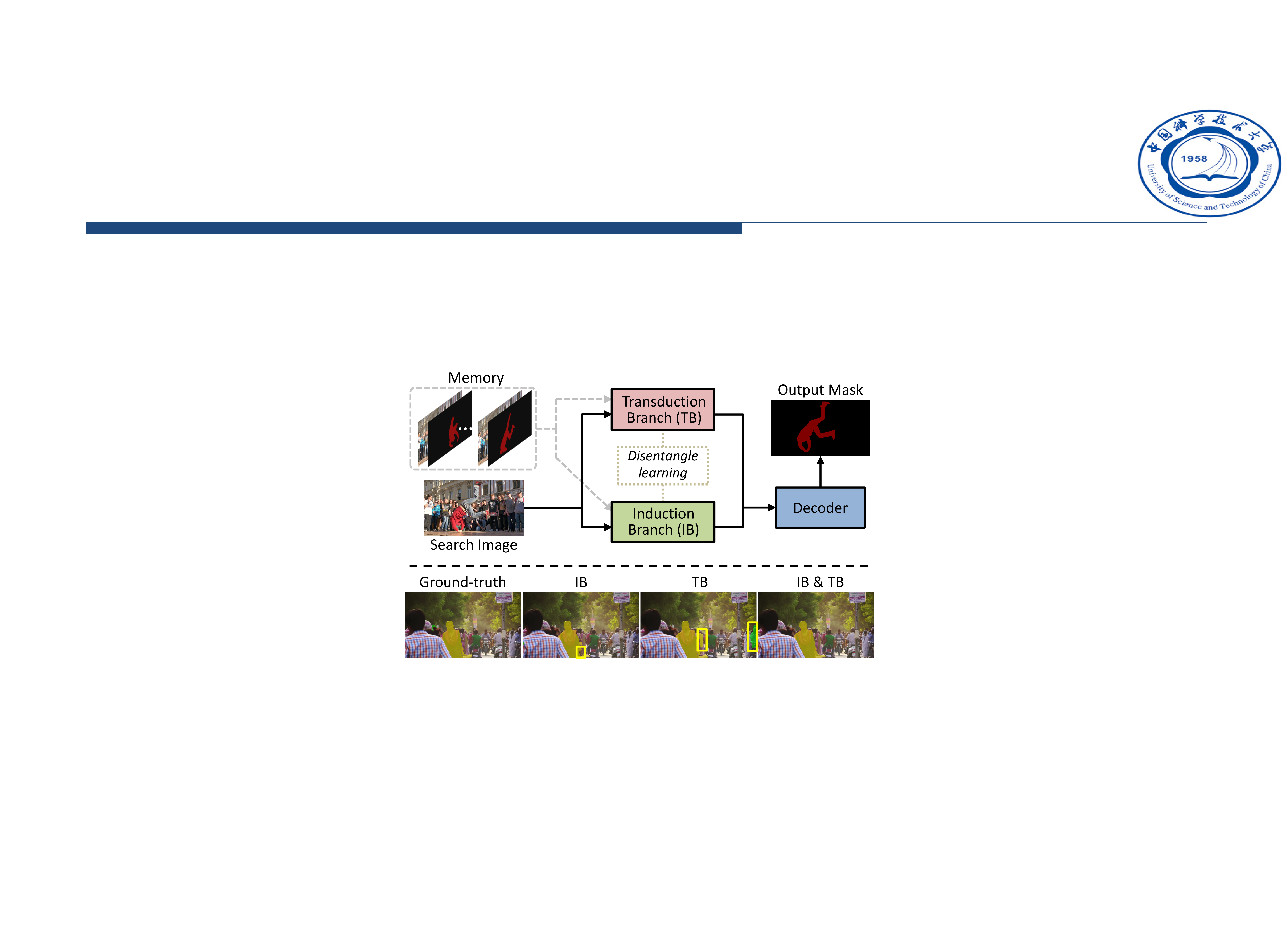}
	\vspace{-2.0em}
	\end{center}
	\caption{An overview of our approach. The transduction branch (TB) aggregates rich spatio-temporal cues from sampled history frames and propagates them to the current frame, and the induction branch (IB) performs online inductive learning to obtain discriminative target information.}
	\label{fig:intro}
	\vspace{-1.0em}
\end{figure}

Semi-supervised video object segmentation (VOS) aims at segmenting the target object in a video sequence with the supervision given in the first frame by a pixel-wise segmentation mask. It has received tremendous attention in recent years for its wide applications. Since the target-specific information is only given in the first frame, and the target may undergo fast-moving and dramatic deformation, how to make full use of the limited information to perform accurate segmentation is thus extremely challenging.

Top-performing methods can be roughly categorized as transductive reasoning and online inductive learning. In transductive formulation, direct reasoning from reference frames (labeled samples) to the current frame (unlabeled sample) is performed to facilitate segmentation.
In recent transductive solutions \cite{contrastrandomwalk2020A,mast2020A,lu2020A,Seoung2019A,Seong2020A,colorvideo2018,ContrastiveWN2021A,CFBI2020A}, feature matching has become the mainstream choice, where pixel-level affinity or distance maps between the current frame and reference frames are obtained to deliver rich historical target information. This specific-to-specific reasoning favorably retains the temporal information with attractive time efficiency. Despite achieving the state-of-the-art performance, it heavily relies on the offline learned feature embeddings for accurate matching, thus suffers limited generalization and discrimination capabilities.

On the other hand, online inductive learning utilizes reference frames to train a target model (general rule), which is then applied to subsequent frames to perform segmentation.
Recently, efficient online discriminative learning \cite{dimp2019A,atom2019A} in visual object tracking has been introduced to the VOS community for its well-acknowledged adaptivity and generalizability. The few-shot learner in \cite{Goutam2020A,frtm2020A} provides superior distractor discrimination capability. 
Nevertheless, this inductive formulation treats reference frames as independent training samples and fails to explore the underlying context. Rich temporal information that resides in the video flow is thus not fully exploited, which has been proven by previous transductive inference attempts \cite{Seoung2019A,Zhang2020A} to be crucial for obtaining spatio-temporal consistent results.

The above analysis indicates that transductive reasoning and online inductive learning are naturally complementary. The former performs better in spatio-temporal dependency modeling but struggles to discriminate similar distractors, while the latter is just the other way around. Although it is intuitive to jointly integrate these two models, how to explore their complementary potentials in a unified framework has been rarely involved. Since they deal with the VOS task via different perspectives, there exist two main challenges for this seemingly straightforward integration. First, most transductive approaches rely on intermediate results as features \cite{Seoung2019A,Seong2020A,li2020A,lu2020A} or distance maps \cite{feelvos2019A,CFBI2020A}, while the online inductive learning directly outputs masks \cite{META2019A,frtm2020A} or intermediate encodings \cite{Goutam2020A}. How to design an appropriate merging strategy to effectively fuse these diverse representations while retaining their complementarity is an open problem.
Second, how to tightly bridge these two different models to avoid redundant computations for efficient online VOS deserves further exploration.

In this work, as shown in Figure \ref{fig:intro}, we propose a novel two-branch architecture to jointly integrate transductive reasoning and online inductive learning within a unified framework for high-performance VOS. 
The transduction branch aggregates rich spatial-temporal information while the induction branch provides superior discrimination capability.
To solve the aforementioned problems and narrow the gap between the two branches, we make several key designs in the proposed framework:
(1) In the transduction branch, we extend the attention mechanism adopted in previous matching-based VOS frameworks \cite{Seoung2019A,Zhang2020A,Seong2020A} into a lightweight transformer \cite{DETR2020A,transformer2017A} architecture, which is carefully designed to facilitate temporal information propagation.
To unify the inputs and outputs of the two branches, we further adopt a two-head label encoder for them to generate mask encodings as the target information carrier for VOS.
(2) We propose the mask encoding decoupling regularization to reduce their redundancy and make the learned target information more differentiable and complementary.
(3) Finally, our two lightweight branches mutually share plentiful blocks such as backbone,  partial label generator, and segmentation decoder, making our framework efficient and end-to-end trainable.
We perform extensive experiments on DAVIS \cite{DAVIS2017} and YouTube-VOS \cite{Xu2018YouTubeVOSAL} benchmarks. Our proposed approach outperforms other state-of-the-art methods with comparable running efficiency.

In summary, we make the following three contributions:
\begin{itemize}[itemsep= 0 pt,topsep = 0 pt, parsep=0 pt]
	\item We propose a novel two-branch architecture to tackle the video object segmentation, which absorbs the merits of both offline learned transductive reasoning and online inductive learning.
	\item For the transduction branch, a lightweight transformer architecture is proposed to conduct spatio-temporal dependency modeling and content propagation.
	To our best knowledge, this is the first attempt to leverage the transformer architecture in VOS.
	\item To bridge the gap between two branches and better exploit their complementary characteristics, we propose to learn disentangled mask encodings.
\end{itemize}

\begin{figure*}
	\begin{center}
	\includegraphics[width=1.0\linewidth]{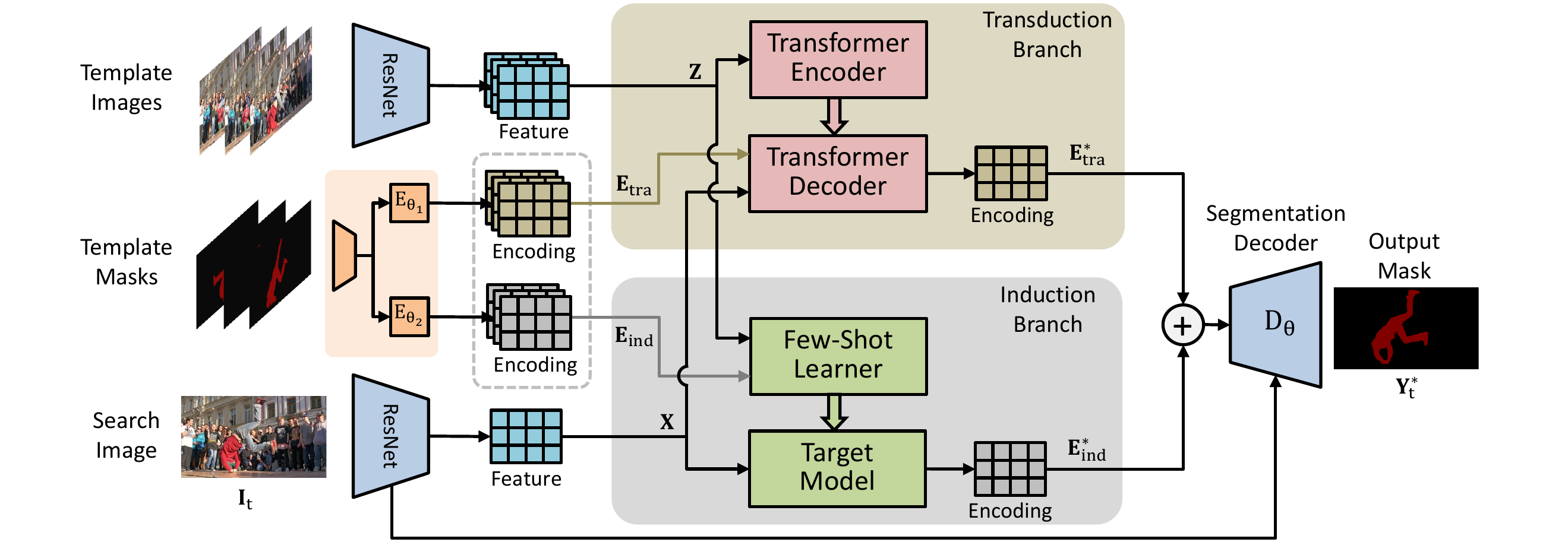}
	\vspace{-2.0em}
	\end{center}
	\caption{The overall pipeline of our approach. It consists of two complementary branches. The lightweight transformer architecture in transduction branch produces fine-grained and spatio-temporal consistent cue while the few-shot learner in induction branch provides discriminative information through online inductive learning. Two branches are integrated by learning disentangled mask encodings.}
	\label{fig:architecture}
	\vspace{-1.0em}
\end{figure*}

\section{Related Work}
\noindent\textbf{Matching based methods.}
Recent methods  \cite{li2020A,lu2020A,Seoung2019A,Seong2020A,PLM2017A,feelvos2019A,ranet2019A,CFBI2020A,Zhang2020A} adopt feature matching for video object segmentation. In these methods, embeddings are firstly obtained through a backbone network for both reference and current frames. Pixel-level comparison is then performed between them to obtain target-specific information for the current frame, which may further be fed into a segmentation decoder to obtain the final result. Among them, STMVOS \cite{Seoung2019A} maintains a memory bank during online inference, and feature matching is realized by applying the non-local cross-attention mechanism, where the memory embeddings are read out to facilitate object segmentation in the current frame. GC \cite{li2020A}, KMN \cite{Seong2020A}, and GraphMemVOS \cite{lu2020A} improve STMVOS in different aspects. TVOS \cite{Zhang2020A} is the first to pose video object segmentation from a transductive inference perspective. In TVOS, a spatio-temporal dependency graph is constructed by adopting a predefined pixel-wise similarity metric. And the graph is utilized to propagate labels from sampled history frames to the current frame.

Different from previous methods \cite{Seoung2019A,Zhang2020A} that adopt isolated attention mechanisms for transductive reasoning, in our transduction branch, we introduce a lightweight yet complete transformer architecture to perform spatio-temporal dependency modeling and target information propagation. And we further learn the intermediate representations suitable for the propagation operation.

\noindent\textbf{Online learning based methods.} 
In many early works \cite{Bao2018A,Cae2017A,static2017A,dyenet2018,premvos2018A,Man2018B,onavos2017A,MoNet2018A},
online fine-tuning is widely used to online introduce the target information. Despite the promising results, this plain inductive learning method is extremely time-consuming, making it unfavorable for many real-time applications. Thus, many efforts \cite{blazingly2018A,TrackPart2018A,d3s2020A,siamask2019A,OSMN2018A,Zhang_luhuchuan_2019A} have been made to avoid first-frame fine-tuning.

In visual object tracking, online discriminative learning  \cite{dimp2019A,eco2017A,atom2019A,prdimp2020A,Wang_2021_Transformer} has received significant attention for its superior performance and high efficiency. In these methods, efficient optimization strategies are adopted to online train convolutional filters, which are applied to subsequent frames to carry out foreground-background classification. In \cite{META2019A}, the closed-form ridge regression \cite{MLCS2019A} is introduced to solve the VOS problem, it online optimizes a parameter matrix that maps features to segmentation masks. In FRTM \cite{frtm2020A}, the online optimization paradigm in ATOM \cite{atom2019A} is revisited and being carefully modified to adapt to the VOS task. By applying Conjugate Gradient and Gauss-Newton algorithms, the few-shot learner in FRTM is capable of learning a powerful target-specific model from the limited number of templates during inference. The learned target model conducts foreground-background classification in a pixel-wise manner, and the obtained low-resolution result is further refined by a segmentation decoder. Later LWL \cite{Goutam2020A} further proposes to learn what the few-shot learner should learn. Different from FRTM, it adopts a label encoder to generate the few-shot label, which contains richer target information than the single-channel segmentation mask. Thanks to the online inductive learning, the capability to discriminate against similar instances is greatly enhanced. But the spatio-temporal consistency of the result may not be well guaranteed.
In this work, we aim to integrate the aforementioned matching-based transductive reasoning and online inductive learning into a unified framework to fully exploit the complementary characteristics between them.

\section{Method}

\subsection{Overall Pipeline} \label{subsection:pipeline}
We propose a new approach for video object segmentation, which consists of two functional branches. The main idea is that matching-based transductive reasoning and online inductive learning are naturally complementary.
The overall pipeline of our approach is illustrated in Figure \ref{fig:architecture}. The first frame and sampled history frames constitute the template images, and the current frame serves as the search image. Firstly, both template and search images are fed into the ResNet-50 \cite{resnet2016A} network to obtain the \texttt{res3} features $\mathbf{Z} \in \mathbb{R}^{N \times H \times W \times C}$ and $\mathbf{X} \in \mathbb{R}^{H \times W \times C}$ respectively, where $N$ is the number of template images. A two-head label encoder is employed to encode the template masks into mask encodings $\mathbf{E}_{\text{tra}} \in \mathbb{R}^{N \times H \times W \times D }$ and $\mathbf{E}_{\text{ind}} \in \mathbb{R}^{N \times H \times W \times D}$ for two parallel branches. After that, the transduction branch takes both $\mathbf{Z}$ and $\mathbf{X}$ as input and propagates $\mathbf{E}_{\text{tra}}$ to search image according to the pixel-level affinity between features. The propagated result is denoted as $\mathbf{E}_{\text{tra}}^* \in \mathbb{R}^{H \times W \times D}$. Meanwhile, the online few-shot learner in the induction branch learns a target model by solving an optimization problem where $\mathbf{Z}$ and $\mathbf{E}_{\text{ind}}$ are treated as training sample pairs. The target model is then applied to $\mathbf{X}$ to obtain $\mathbf{E}_{\text{ind}}^* \in \mathbb{R}^{H \times W \times D}$ of the search image, which contains discriminative target information. Finally, the obtained mask encodings from both branches along with search features from different backbone layers are incorporated and fed into the segmentation decoder to predict the final result. Note that two branches are integrated in a complementary manner by learning disentangled mask encodings, which will be discussed in Section \ref{subsection:ldl}.

\subsection{Transduction Branch} \label{subsection:mpb}
 
\begin{figure}[t]
	\begin{center}
	\includegraphics[width=1.0\linewidth]{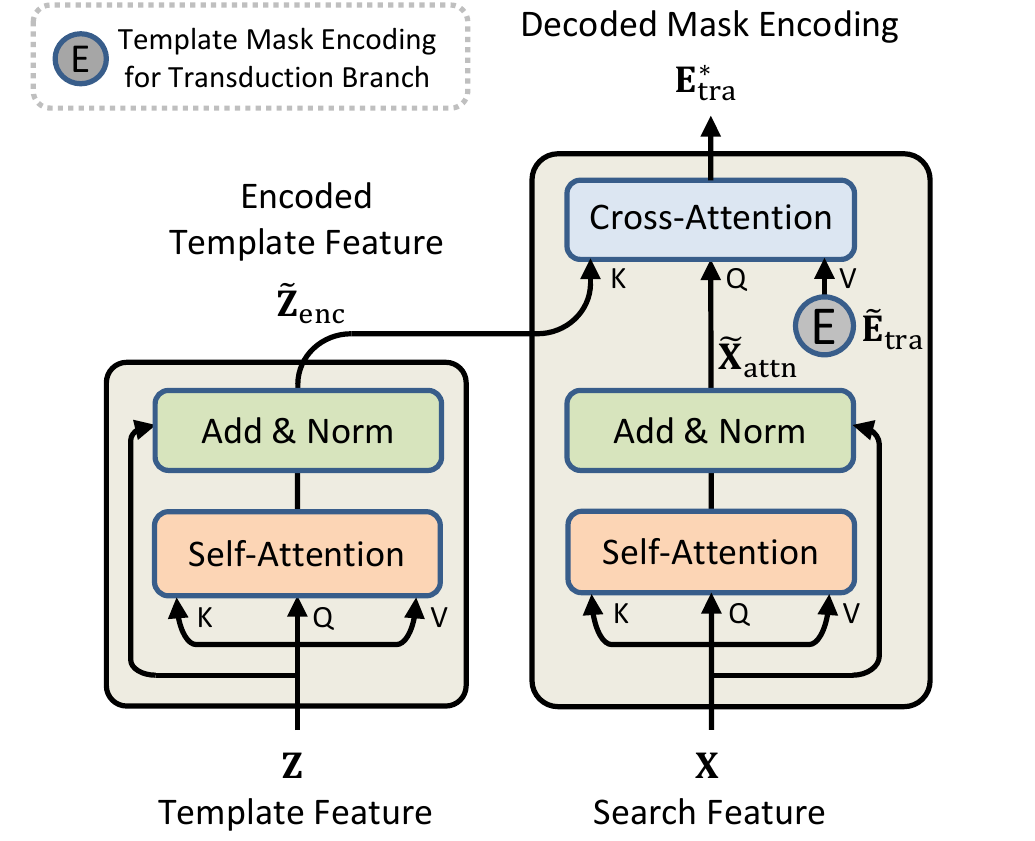}
	\vspace{-2.0em}
	\end{center}
	\caption{An overview of our lightweight transformer architecture, which is adopted in the transduction branch. It is carefully designed to provides fine-grained and temporally consistent target information propagation.}
	\label{fig:transformer}
	\vspace{-0.5em}
\end{figure}

As shown in Figure \ref{fig:transformer}, in the transduction branch, a lightweight transformer \cite{transformer2017A} architecture is introduced to perform spatio-temporal information transduction. The attention mechanism, which is the most important component in transformer, has a strong capability in non-local dependency modeling. It transforms value $\mathbf{V} \in \mathbb{R}^{n_k \times d_v}$ according to the dot-product similarity between query $\mathbf{Q} \in \mathbb{R}^{n_q \times d_k}$ and key $\mathbf{K} \in \mathbb{R}^{n_k \times d_k}$. Here in our approach, the attention mechanism is slightly modified to better adapt to the VOS task. Firstly, query and key are normalized along the channel dimension before the dot-product operation. Then, the intermediate result $\bar{\mathbf{Q}}\bar{\mathbf{K}}^\top \in \mathbb{R}^{n_q \times n_k}$ is rescaled to obtain a suitable softmax distribution \cite{chen2020simple,hinton2015distilling}. The above computing process can be formulated as follows:
\begin{equation}\label{tau}
	\text{Attention}(\mathbf{Q}, \mathbf{K}, \mathbf{V}) = \text{Softmax}
	\left( \frac{ \bar{\mathbf{Q}}\bar{\mathbf{K}}^\top}{\tau}
	\right)\mathbf{V},
\end{equation}
where $\bar{\cdot}$ denotes $\ell_{2}$ normalization along channel dimension and the denominator $\tau$ is the scaling factor.

\noindent\textbf{Transformer encoder}.
The transformer encoder takes template feature $\mathbf{Z} \in \mathbb{R}^{N \times H \times W \times C}$ as input, which is further flattened into $\widetilde{\mathbf{Z}} \in \mathbb{R}^{NHW \times C }$ for subsequent matrix operations. In self-attention layer, the query and key are obtained by applying a linear transformation $\phi(\cdot)$ to the flattened feature $\widetilde{\mathbf{Z}}$ whose channel dimension is reduced from $C$ to $C/4$. And the attention value $\mathbf{A}_{\widetilde{\mathbf{Z}}} \in \mathbb{R}^{NHW \times C}$ is computed according to Eq. \eqref{tau} as follows:
\begin{equation}
	\mathbf{A}_{\widetilde{\mathbf{Z}}} = 
	\text{Attention}
	\left( \phi(\widetilde{\mathbf{Z}}),\phi(\widetilde{\mathbf{Z}}), \widetilde{\mathbf{Z}}
	\right).
\end{equation}
This attention value, as a residual term, is added to the original template feature $\widetilde{\mathbf{Z}}$, and the result is further fed into the Instance Normalization \cite{instnorm2016A} layer to obtain the encoded template feature $\widetilde{\mathbf{Z}}_{\text{enc}} \in \mathbb{R}^{NHW \times C}$ as follows:
\begin{equation}
	\widetilde{\mathbf{Z}}_{\text{enc}} = \text{Instance Norm}\left(\mathbf{A}_{\widetilde{\mathbf{Z}}} + \widetilde{\mathbf{Z}} \right).
\end{equation}
The transformer encoder enables template features to mutually reinforce to be more compact and representative, thus suitable for subsequent feature matching procedures performed in the transformer decoder.

\noindent\textbf{Transformer decoder}.
The transformer decoder consists of a self-attention layer and a cross-attention layer. Firstly, the self-attention layer processes search feature $\mathbf{X} \in \mathbb{R}^{H \times W \times C}$ in a similar way to the transformer encoder, \textit{i.e.}, a residual attention term is obtained and being merged to the original search feature as follows:
\begin{equation}
	\begin{aligned}
		&\mathbf{A}_{\widetilde{\mathbf{X}}} = 
		\text{Attention}
		\left( \phi(\widetilde{\mathbf{X}}),\phi(\widetilde{\mathbf{X}}), \widetilde{\mathbf{X}}
		\right),  \\
		&\widetilde{\mathbf{X}}_{\text{attn}} = \text{Instance Norm}\left( \mathbf{A}_{\widetilde{\mathbf{X}}} + \widetilde{\mathbf{X}} \right),
	\end{aligned}
\end{equation}
where $\widetilde{\mathbf{X}} \in \mathbb{R}^{HW \times C}$ is the flattened search feature, $\mathbf{A}_{\widetilde{\mathbf{X}}} \in \mathbb{R}^{HW \times C}$ and $\widetilde{\mathbf{X}}_{\text{attn}} \in \mathbb{R}^{HW \times C}$ denote the attention value and output of the self-attention layer, respectively. Then, the cross-attention layer, which is the most important component in our lightweight transformer architecture, propagates rich temporal information according to the pixel level correspondence between the search image and template images. It takes $\widetilde{\mathbf{X}}_{\text{attn}} \in \mathbb{R}^{HW \times C}$  and the encoded template feature $\widetilde{\mathbf{Z}}_{\text{enc}} \in \mathbb{R}^{NHW \times C}$ as inputs to generate query $\psi(\widetilde{\mathbf{X}}_{\text{attn}})$ and key $\psi(\widetilde{\mathbf{Z}}_{\text{enc}})$ respectively, and transforms template mask encoding $\mathbf{E}_{\text{tra}} \in \mathbb{R}^{N \times H \times W \times D}$ to search image according to the similarity between query and key as follows:
\begin{equation}
	\widetilde{\mathbf{E}}_{\text{tra}}^{*} =
	\text{Attention}
	\left( \psi(\widetilde{\mathbf{X}}_{\text{attn}}),\psi(\widetilde{\mathbf{Z}}_{\text{enc}}), \widetilde{\mathbf{E}}_{\text{tra}}
	\right),  \\
\end{equation}
where $\mathbf{E}_{\text{tra}}$ is flattened into $\widetilde{\mathbf{E}}_{\text{tra}} \in \mathbb{R}^{NHW \times D}$ before being used as the value. And $\widetilde{\mathbf{E}}_{\text{tra}}^{*} \in \mathbb{R}^{HW \times D}$ is reshaped back to obtain the decoded mask encoding $\mathbf{E}_{\text{tra}}^{*} \in \mathbb{R}^{H \times W \times D}$. 

\subsection{Induction Branch}\label{subsection:db}
The transduction branch provides fine-grained and temporally consistent mask encoding propagation, but its performance largely relies on the representation capability of the offline trained feature embeddings. Due to the absence of online adaptation, it does not perform well when encountering novel targets, and it is also difficult to discriminate similar instances. To make up for these shortcomings, in the induction branch, we adopt the few-shot learner proposed in LWL \cite{Goutam2020A} to perform online inductive learning. Taking template feature $\mathbf{Z}$ and mask encoding $\mathbf{E}_{\text{ind}}$ as training sample pairs, it online optimizes the kernel of a convolutional layer $T_{\boldsymbol{\omega}}: \mathbb{R}^{H \times W \times C} \to \mathbb{R}^{H \times W \times D}$ by minimizing the following squared error:
\begin{equation}\label{equation:minsquare}
	L(\boldsymbol{\omega}) = \frac{1}{2} \sum\limits_{i=1}^{N}\Vert\mathbf{W}_i \cdot \big( T_{\boldsymbol{\omega}}(\mathbf{Z}_i)-\mathbf{E}_{\text{ind},i}\big)\Vert^2 + \frac{\lambda}{2}\Vert \boldsymbol{\omega} \Vert^2,
\end{equation}
where $\boldsymbol{\omega}$ is the kernel to be optimized, $\mathbf{W}$ represents the element-wise importance weights generated from labels (as with the mask encodings), $i$ is the index of training samples, and $\lambda$ is a learned regularization term. The steepest descent method is applied to iteratively minimize the squared error instead of the direct closed-form solution \cite{MLCS2019A}, as the latter requires time-consuming matrix inversion operations which are harmful to the running speed. For the detailed derivation of the steepest descent method please refer to \cite{Goutam2020A}. Note that the entire optimization process is fully differentiable, so it can be offline trained together with the rest part of the network in an end-to-end manner.

This online optimized inductive target model, \textit{i.e.}, the convolutional kernel $\boldsymbol{\omega} \in \mathbb{R}^{K \times K \times C \times D}$, has excellent discrimination capability. It maps the search feature $\mathbf{X}$ into a $D$-dimensional target-aware mask encoding $\mathbf{E}_{\text{ind}}^*$, which greatly compensates for the output of the transduction branch.

\subsection{Disentangled Mask Encodings}\label{subsection:ldl}
As stated above, both branches produce mask encodings with rich target specific information. These mask encodings are element-wisely added together in our approach. And the result, along with the search features from different backbone layers, are processed by the segmentation decoder $D_{\theta}$ to generate the mask prediction $\mathbf{Y}_{t}^*$ as follows:
\begin{equation}
	\mathbf{Y}_{t}^* = D_{\theta}\left(\mathbf{E}_{\text{tra}}^{*} + \mathbf{E}_{\text{ind}}^{*}, \mathbf{X}_t^{[1,2,3,4]}\right),
\end{equation}
where $\mathbf{X}_t^{[1,2,3,4]}$ is a simplified denotation of search features from different backbone layers (layer 1 to 4).

\begin{figure}[tbp]
	\begin{center}
	\includegraphics[width=1.0\linewidth]{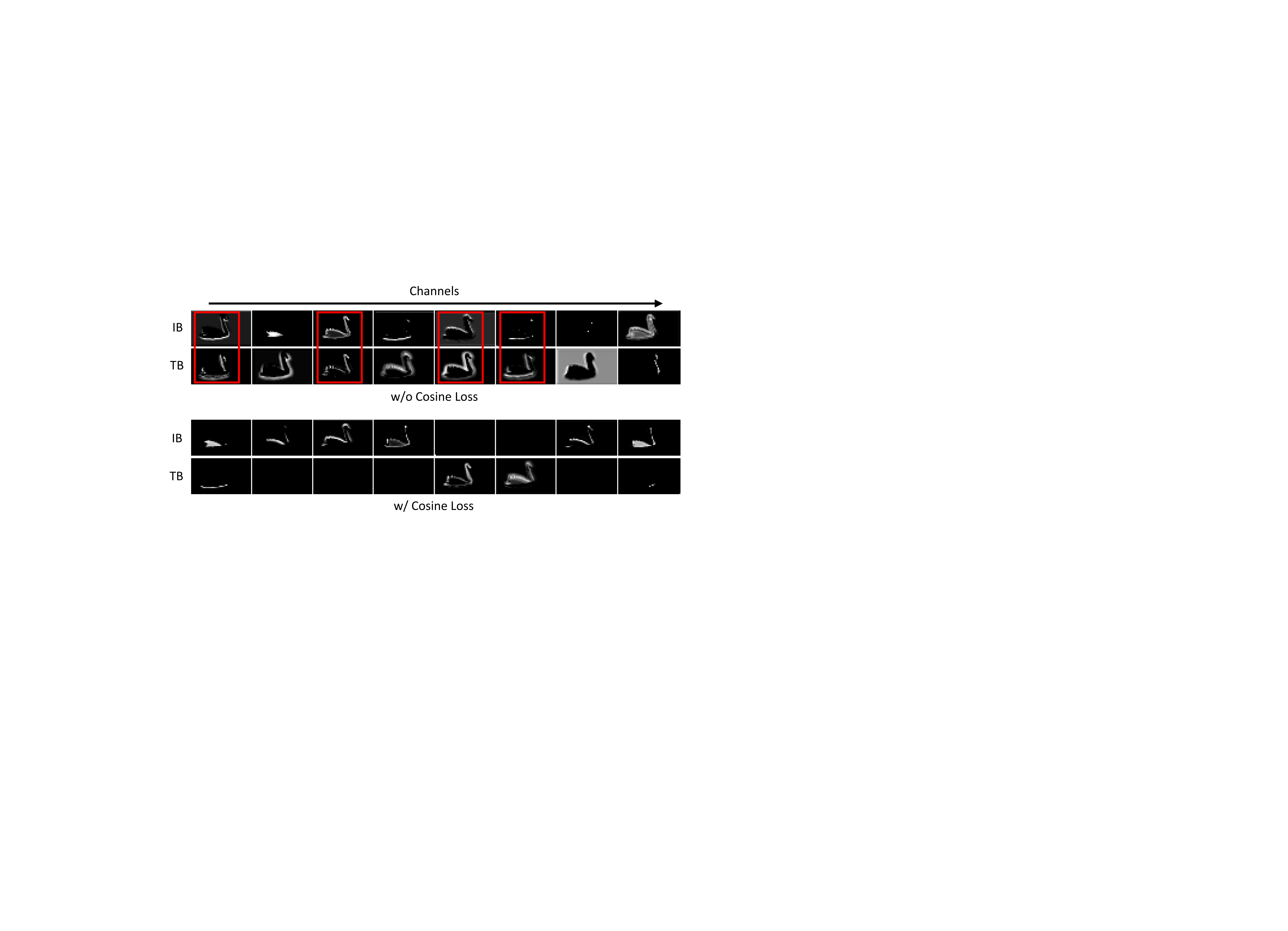}
	\vspace{-2.0em}
	\end{center}
	\caption{Visualization of the generated mask encodings. IB and TB denote induction branch and transduction branch, respectively.
		We can find that different branches learn diverse mask encodings.
		However, without any constraints, the generated encodings have plenty of information redundancy (marked in red boxes), which is alleviated by adopting the cosine loss in our approach.}
	\label{fig:mask_enc_show}
	\vspace{-0.5em}
\end{figure}

Since the two branches show diverse characteristics in dealing with the VOS task, the mask encoding suitable for each of them should be different. To this end, we propose a two-head label encoder to learn the intermediate expressions suitable for each branch. Furthermore, to reduce their redundancy and make the learned target information more differentiable and complementary, we decouple the generated mask encodings by minimizing their similarity. Specifically, we adopt the widely used cosine similarity in our approach. Given two vectors $a$ and $b$, the cosine similarity can be calculated by $\operatorname{cos}(a,b)=\frac{a \cdot b}{\|a\|\cdot\|b\|}$. The adopted regularization loss is defined as follows:
\begin{equation}\label{equation:losscos}
	\mathcal{L}_{\text{cos}}^t = 
	\left\{
	\begin{array}{lll}
		\cos (\operatorname{vec}(E_{\theta_1}(\mathbf{Y}_0)), \operatorname{vec}(E_{\theta_2}(\mathbf{Y}_0)) ) & t=0\\
		\cos (\operatorname{vec}(E_{\theta_1}(\mathbf{Y}_t^*)), \operatorname{vec}(E_{\theta_2}(\mathbf{Y}_t^*)) ) & t > 0\\
	\end{array}
	\right.,
\end{equation}
where $\operatorname{vec}(\cdot)$ is the vectorization operator that flattens the mask encodings into one-dimensional vectors, $\mathbf{Y}_0$ is the ground-truth of $\mathbf{I}_0$ and $\mathbf{Y}_t^*$ is the predicted mask of $\mathbf{I}_t$, $\theta_1$ and $\theta_2$ denote different parameters of the two-head label encoder. Note that since each head ends with a ReLU layer, the cosine similarity will not be negative.
As shown in Figure \ref{fig:mask_enc_show}, the two-head label encoder generates diverse mask encodings for each branch. But there still exists plenty of information redundancy in between. And if we further adopt the regularization loss defined in Eq. \eqref{equation:losscos}, the generated mask encodings are well decoupled along the channel dimension. This ensures that after element-wise addition, the two mask encodings do not mutually disturb. So that the segmentation decoder can make use of the complementary information provided by the two branches to generate the final segmentation result.

\subsection{Online Inference}\label{subsection:inference}
During online inference, a memory bank $\mathbf{M}$ is maintained to memorize the sampled history images and corresponding segmentation masks. Given a test video sequence $\mathcal{S} = \{\mathbf{I}_t\}_{t=0}^{N_{\text{test}}}$ with the initial segmentation mask $\mathbf{Y}_0$ of the first frame $\mathbf{I}_0$, we first initialize the memory bank with $\mathbf{M}_0 = \{(\mathbf{X}_0, \mathbf{Y}_0)\}$, where $\mathbf{X}_0 = F_{\theta}(\mathbf{I}_0)$ is the \texttt{res3} feature extracted from $\mathbf{I}_0$ by the backbone network $F_{\theta}$. Then the initialized memory bank is used to perform object segmentation on subsequent frames as stated in Section \ref{subsection:pipeline}. To better exploit the temporal information and adapt to the appearance changes in the scene, we update the memory bank with the most recently processed frames. Specifically, the new template is sampled every $T =5$ frame and added to the memory bank along with the predicted segmentation mask. And except for the first frame, we remove the oldest sample to ensure that the size of the memory bank does not exceed $N_{\text{max}}=20$. For multiple objects, our approach processes each of them independently and merges the predicted masks using the soft-aggregation operation \cite{rgmp2018A}. Note that the merging step is performed only during online inference.

\subsection{Offline Training}\label{subsection:training}
The whole network in our approach is end-to-end trained using generated mini-sequences $\mathcal{S}_{\text{train}} = \{(\mathbf{I}_t, \mathbf{Y}_t)\}_{t=0}^{N_{\text{train}}-1}$, which are randomly sampled from densely annotated video segments.
During offline training, our network processes mini-sequences in a similar way to the online inference stage. To fully exploit the generated mini-sequences, the memory bank is updated every frame during offline training. We adopt two loss functions to supervise the learning of our network, namely segmentation loss and cosine similarity loss. The segmentation loss is used to supervise the generated mask predictions, which is computed as follows:
\begin{equation}
	\mathcal{L}_{\text{seg}}^t =\mathcal{L}_{\text{lov}} \left(\mathbf{Y}_t^*, \mathbf{Y}_t\right),
\end{equation}
where $\mathbf{Y}_t^*$ is the predicted segmentation mask of $\mathbf{I}_t$, and $\mathcal{L}_{\text{lov}}$ is the Lovasz segmentation loss \cite{Lovasz2018A}. The final loss $\mathcal{L}_{\text{final}}$ is the weighted sum of the segmentation loss $\mathcal{L}_{\text{seg}}^t$ and the aforementioned cosine similarity loss $\mathcal{L}_{\text{cos}}^t$, as follows:
\begin{equation}
	\mathcal{L}_{\text{final}} =  
	\frac{1}{N_{\text{train}}-1} \sum\limits_{t=1}^{N_{\text{train}}-1} \mathcal{L}_{\text{seg}}^t
	+
	\frac{\lambda}{N_{\text{train}}} \sum\limits_{t=0}^{N_{\text{train}}-1}
	\mathcal{L}_{\text{cos}}^t
	,
\end{equation}
where the hyperparameter $\lambda$ is set to $0.01$.

\section{Experiments}
We evaluate the proposed approach on DAVIS 2017 \cite{DAVIS2017} and YouTube-VOS \cite{Xu2018YouTubeVOSAL} datasets. For the DAVIS benchmark, we follow its standard protocol, where the $\mathcal{J}$ score measures the region similarity, the $\mathcal{F}$ score indicates boundary accuracy, and $\mathcal{J} \& \mathcal{F}$ is the mean of them. For comparison on YouTube-VOS datasets, $\mathcal{J}$ and $\mathcal{F}$ scores are reported on both the training (seen) categories and the unseen categories, and the overall score is their average. All results are obtained through the official evaluation toolkit (for DAVIS) or evaluation server (for YouTube-VOS).

\subsection{Implementation Details}
The backbone feature extractor used in our approach is ResNet-50 \cite{resnet2016A}, which is initialized with the Mask R-CNN \cite{maskrcnn2017A} weights. In both branches, an additional convolutional block is adopted to reduce the channel dimension of backbone \texttt{res3} feature from $1024$ to $512$. The scaling factor $\tau$ in Eq. \eqref{tau} is set to $1/30$. For the few-shot learner in the induction branch, we follow the settings used in LWL \cite{Goutam2020A}. The adopted two-head label encoder generates mask encodings with channel dimension $D=16$.

Template and search images are cropped from original frames, which are $5$ times the previously estimated target size (no larger than the original frames). And the cropped patches are further resized to $832 \times 480$.

Our network is trained on the train split of YouTube-VOS \cite{Xu2018YouTubeVOSAL} and DAVIS \cite{DAVIS2016} datasets. We sample $N_{\text{train}}=4$ frames from a video segment of length $N_{\text{train}}^{'}=100$ to generate the mini-sequences, where the random flipping, rotation, and scaling are adopted for data augmentation. The whole training process contains 180k iterations with a batch size of 20. The ADAM \cite{ADAM2015A} optimizer is adopted and the initial learning rate is set to 0.01, which is further reduced by a factor of 5 after 40k, 80k, 115k, and 165k iterations. The backbone weights are fixed in the first 90k iterations and then being optimized together in the rest 90k ones. It takes about 96 hours on 8 Nvidia GTX 1080Ti GPUs to finish the offline training process. During online inference, our approach operates at about 8 FPS on single object sequences.
Code and pre-trained models will be made publicly available.

\begin{table}[tbp]
	\footnotesize
	\vspace{0.4em}
	\caption{Ablation study for branch complementarity. TB and IB denote the transduction branch and  induction branch, respectively. t/s denotes second per frame. The performance is evaluated on the YouTube-VOS 2018 \cite{Xu2018YouTubeVOSAL} validation set in terms of mean Jaccard ($\mathcal{J}$) and	boundary ($\mathcal{F}$) scores on both seen and unseen categories.}
	\vspace{-1.0em}
	\begin{center}
		\begin{tabular*}{\hsize}{@{}@{\extracolsep{\fill}}lccccccc@{}}
			\toprule[1.0pt]
			TB & IB & $\mathcal{J}_{\text{seen}}$ & $\mathcal{F}_{\text{seen}}$ & $\mathcal{J}_{\text{unseen}}$ & $\mathcal{F}_{\text{unseen}}$ & Overall & t/s\\
			\midrule
			\checkmark &  & 81.2 & 85.4 & 75.1 & 83.2 & 81.2 & 0.22\\
			& \checkmark & 80.4 & 84.9 & 76.4 & 84.4 & 81.5 & 0.15\\
			\checkmark & \checkmark & \textbf{81.5} & \textbf{85.9} & \textbf{78.7} & \textbf{86.5} & \textbf{83.1} & 0.25\\
			\bottomrule[1.0pt]
		\end{tabular*}
	\end{center}
	\label{table:branch-comp}
	\vspace{-1.5em}
\end{table}

\begin{table}[tbp]
	\footnotesize
	\caption{Ablation study for disentangled mask encodings. For the single-head label encoder, two branches share the same mask encoding. The performance is evaluated on the YouTube-VOS 2018 \cite{Xu2018YouTubeVOSAL} validation set in terms of mean Jaccard ($\mathcal{J}$) and boundary ($\mathcal{F}$) scores on both seen and unseen categories.}
	\vspace{-1.0em}
	\begin{center}
		\setlength{\tabcolsep}{0mm}{
			\begin{tabular*}{\hsize}{@{}@{\extracolsep{\fill}}cccccccc@{}}
				\toprule[1.0pt]
				\multirow{2}*{Version} & Label & Cosine & \multirow{2}*{$\mathcal{J}_{\text{seen}}$} & \multirow{2}*{$\mathcal{F}_{\text{seen}}$} &
				\multirow{2}*{$\mathcal{J}_{\text{unseen}}$} &
				\multirow{2}*{$\mathcal{F}_{\text{unseen}}$} &
				\multirow{2}*{Overall}\\
				& Encoder & Loss & & & & & \\
				\midrule
				(1) & single-head & & 80.7 & 85.3 & 76.8 & 84.5 & 81.8\\
				(2) & two-head & & 81.1 & 85.6 & 77.6 & 85.3 & 82.4\\
				(3) & two-head & \checkmark & \textbf{81.5} & \textbf{85.9} & \textbf{78.7} & \textbf{86.5} & \textbf{83.1}\\
				\bottomrule[1.0pt]
			\end{tabular*}
		}
	\end{center}
	\label{table:disentangle}
	\vspace{-2.0em}
\end{table}

\subsection{Ablation Study}
To validate the effectiveness of the key components in our proposed method, we perform two comparative experiments on the YouTube-VOS 2018 \cite{Xu2018YouTubeVOSAL} validation set.

\begin{figure}[htbp]
	\begin{center}
	\includegraphics[width=1.0\linewidth]{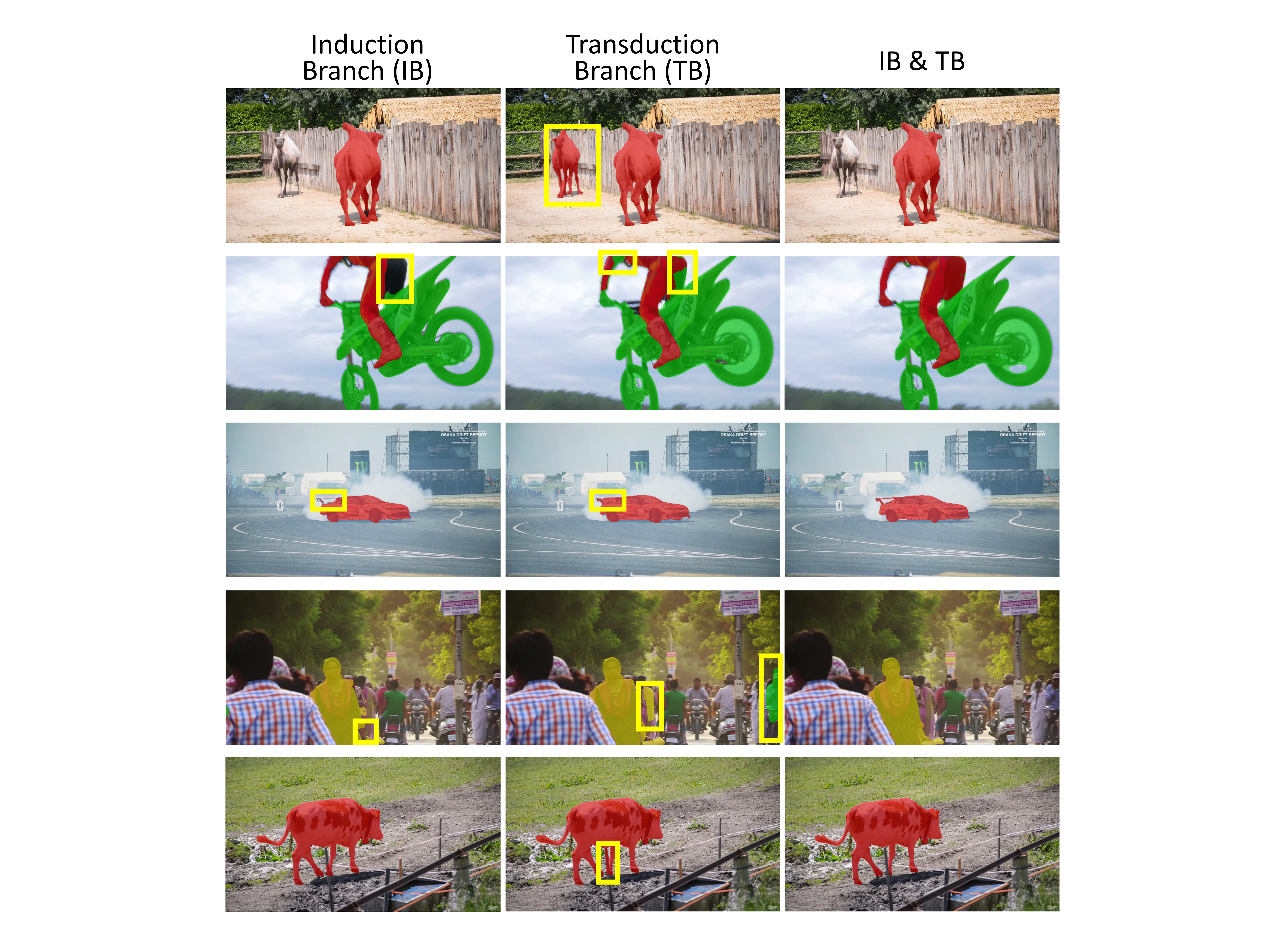}
	\vspace{-2.0em}
	\end{center}
	\caption{Qualitative comparisons on DAVIS 2017 validation set.
		Noticeable failures are marked with yellow boxes. By jointly exploring induction branch (IB) and transduction branch (TB), our method shows superior VOS accuracy. Best view in zoom in.}
	\label{fig:qualitative_comp}
	\vspace{-1.0em}
\end{figure}

\noindent\textbf{Branch complementarity.}
We first perform an ablation study to demonstrate the complementarity of the two branches in our approach. The experimental results are presented in Table \ref{table:branch-comp}. The overall performance degrades from $83.1\%$ to $81.2\%$ when the transduction branch is applied alone. And we can further discover that the performance degradation is mainly reflected on the unseen categories, where $\mathcal{J}_{\text{unseen}}$ and $\mathcal{F}_{\text{unseen}}$ are dropped from $78.7 \%$ and $86.5\%$ to $75.1\%$ and $83.2\%$, respectively. This proves the insufficient generalization capability of matching-based approaches to a certain extent. And we further report the result of LWL \cite{Goutam2020A} as the performance of applying induction branch alone. Compared with the full version, LWL \cite{Goutam2020A} has decreased results in all performance metrics, with an overall score of $81.5\%$. The above results show that there does exist a strong complementary relationship between the two branches of our approach. Qualitative comparisons on DAVIS 2017 validation set are shown in Figure \ref{fig:qualitative_comp}.

\noindent\textbf{Disentangled mask encodings.}
We then conduct several ablation experiments to verify the effectiveness of our proposed disentangled mask encodings for exploiting the complementarity of the two branches. Specifically, we set up three mask encoding generation strategies as follows:
\begin{enumerate}[itemsep= 0 pt,topsep = -1 pt, parsep=0 pt]
	\item[(1)] We first adopt a single-head label encoder for the mask encoding generation, \textit{i.e.}, both branches adopt the same mask encoding.
	\item[(2)] We next replace the above-mentioned single-head label encoder with a two-head one. In this setting, mask encoding for each branch is independently generated.
	\item[(3)] Based on (b), in this version, we further introduce the cosine similarity loss proposed in Section \ref{subsection:ldl} to force the generated mask encodings to be disentangled.
\end{enumerate}
As shown in Table \ref{table:disentangle}, directly applying a single-head label encoder achieves an overall score of $81.8\%$. If we replace it with a two-head one, the overall performance is improved from $81.8\%$ to $82.4\%$. This indicates that the intermediate representations of masks suitable for the two branches are different. And if we further apply the cosine similarity loss to force the generated mask encodings to be disentangled, the performance can be further improved to $83.1\%$.

\subsection{Comparison with State-of-the-art Methods}

\begin{figure*}[htbp]
	\begin{center}
	\includegraphics[width=1.0\linewidth]{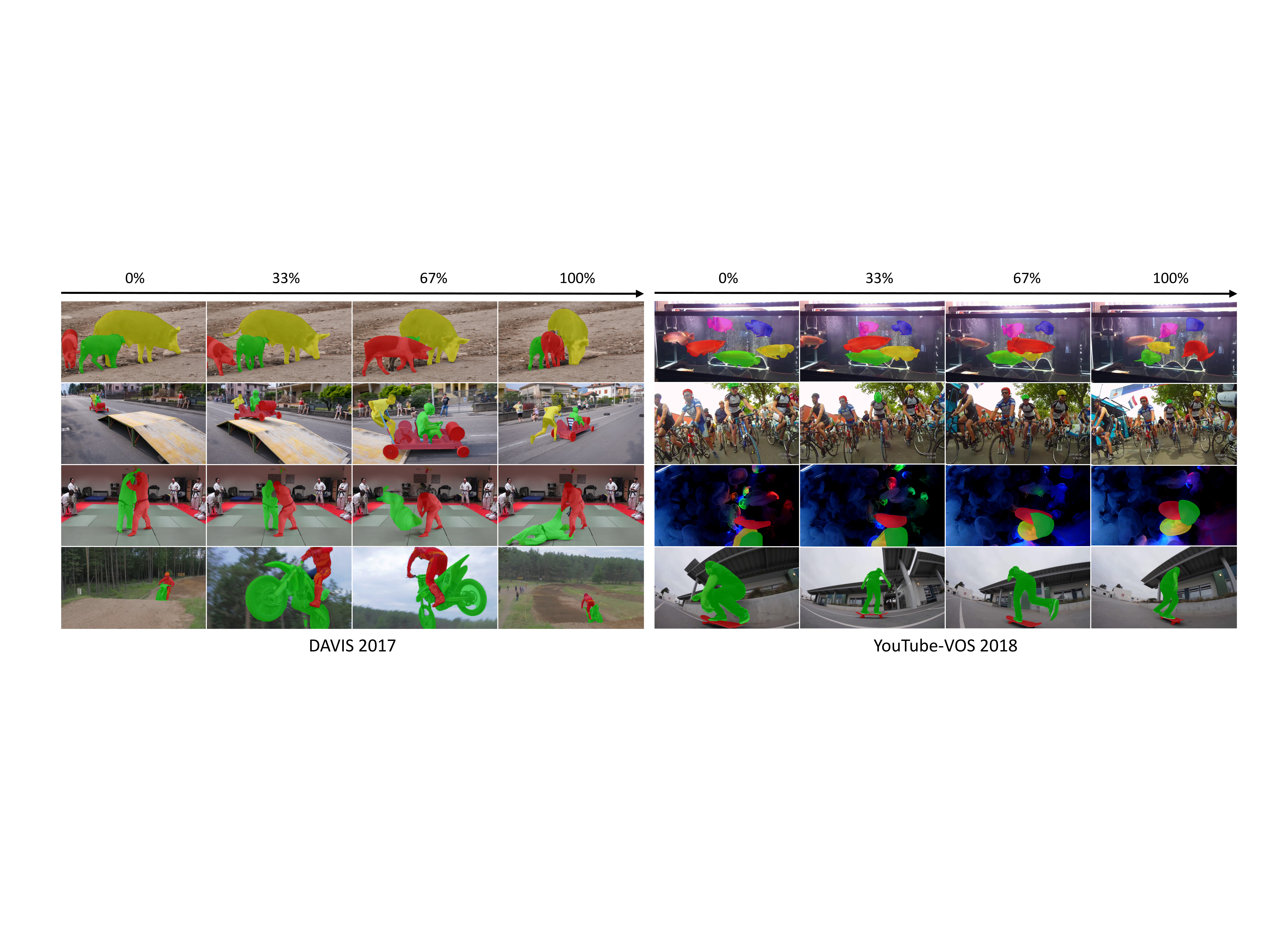}
	\vspace{-2.0em}
	\end{center}
	\caption{Qualitative result of our approach on DAVIS \cite{DAVIS2017} and YouTube-VOS 2018 \cite{Xu2018YouTubeVOSAL} validation sets.
	Our approach shows superior segmentation accuracy on both of them. In the first row, the pig in green and fish in yellow undergo severe occlusion. In the second row, similar distractors and cluttered background exist in the scene. The third and fourth rows are scenes with changes in appearance and perspective, respectively. Our approach successfully handles all these challenging scenarios.}
	\label{fig:qualitative}
	\vspace{-1.0em}
\end{figure*}

We compare our approach with previous state-of-the-art methods on several benchmarks including DAVIS 2017 \cite{DAVIS2017}, YouTube-VOS 2018, and YouTube-VOS 2019 \cite{Xu2018YouTubeVOSAL}. In Figure \ref{fig:qualitative}, we show some qualitative results in many challenging scenarios, such as occlusion, similar distractors, and appearance changes. Our \underline{JO}int \underline{IN}ductive and \underline{T}ransductive learning-based method is denoted as \textbf{JOINT}.


\begin{table}[htbp]
	\footnotesize
	\caption{State-of-the-art comparation on the YouTube-VOS \cite{Xu2018YouTubeVOSAL} validation datasets. S denotes using synthetic data for offline training, FT denotes online fine-tuning. Our approach has superior generalization capability for unseen categories and outperforms all previous methods by considerable margins on both versions.}
	\vspace{-1.0em}
	\begin{center}
		\setlength{\tabcolsep}{0mm}{
		\begin{tabular*}{\hsize}{@{}@{\extracolsep{\fill}}lccccccc@{}}
			\toprule[1.0pt]
			\multicolumn{8}{c}{\emph{Validation 2018 Split}} \\
			\midrule
			Methods & S & FT & $\mathcal{J}_{\text{seen}}$ & $\mathcal{F}_{\text{seen}}$ & $\mathcal{J}_{\text{unseen}}$ & $\mathcal{F}_{\text{unseen}}$ & Overall\\
			\midrule
			OnAVOS  \cite{onavos2017A} & - & \checkmark & 60.1 & 62.7 & 46.1 & 51.7 & 55.2\\
			OSVOS  \cite{Cae2017A} & - & \checkmark & 59.8 & 60.5 & 54.2 & 60.7 & 58.8\\
			PReMVOS  \cite{premvos2018A} & \checkmark & \checkmark & 71.4 & 75.9 & 56.5 & 63.7 & 66.9\\
			SiamRCNN  \cite{siamrcnn2019A} & - & \checkmark & 73.5 & - & 66.2 & - & 73.2\\
			STMVOS  \cite{Seoung2019A} & \checkmark & - & 79.7 & 84.2 & 72.8 & 80.9 & 79.4\\
			EGMN  \cite{lu2020A} & \checkmark & - & 80.7 & 85.1 & 74.0 & 80.9 & 80.2\\
			KMNVOS  \cite{Seong2020A} & \checkmark & - & 81.4 & 85.6 & 75.3 & 83.3 & 81.4\\
			\midrule
			S2S  \cite{seq2seq2018A} & - & - & 71.0 & 70.0 & 55.5 & 61.2 & 64.4\\
			AGAME  \cite{agame2019A} & - & - & 67.8 & 69.5 & 60.8 & 66.2 & 66.1\\
			CFBI  \cite{CFBI2020A} & - & - & 81.1 & 85.8 & 75.3 & 83.4 & 81.4\\
			LWL  \cite{Goutam2020A}  & - & - &  80.4 & 84.9 & 76.4 & 84.4 & 81.5\\
			CFBI$^{MS}$  \cite{CFBI2020A} & - & - & \textbf{82.2} & \textbf{86.8} & 76.9 & 85.0 & 82.7\\
			\textbf{JOINT (Ours)} & - & - & 81.5 & 85.9 & \textbf{78.7} & \textbf{86.5}& \textbf{83.1}\\
			\midrule[1.0pt]
			\multicolumn{8}{c}{\emph{Validation 2019 Split}} \\
			\midrule
			Methods & S & FT & $\mathcal{J}_{\text{seen}}$ & $\mathcal{F}_{\text{seen}}$ & $\mathcal{J}_{\text{unseen}}$ & $\mathcal{F}_{\text{unseen}}$ & Overall\\
			\midrule
			STMVOS  \cite{Seoung2019A} & \checkmark & - & 79.6 & 83.6 & 73.0 & 80.6 & 79.2\\
			\midrule
			LWL  \cite{Goutam2020A} & - & - & 79.6 & 83.8 & 76.4 & 84.2 & 81.0\\
			CFBI  \cite{CFBI2020A} & - & - & 80.6 & 85.1 & 75.2 & 83.0 & 81.0\\
			CFBI$^{MS}$  \cite{CFBI2020A} & - & - & \textbf{81.8} & \textbf{86.1} & 76.9 & 84.8 & 82.4\\
			\textbf{JOINT (Ours)} & - & - & 80.8 & 84.8 & \textbf{79.0} & \textbf{86.6} & \textbf{82.8}\\
			\bottomrule[1.0pt]
		\end{tabular*}
		}
	\end{center}
	\label{table:yt-val}
	\vspace{-2.5em}
\end{table}

\begin{table}[htbp]
	\footnotesize
	\vspace{0.6em}
	\caption{State-of-the-art comparation on the DAVIS 2017 \cite{DAVIS2017} validation dataset. S denotes using synthetic data for offline training, FT denotes online fine-tuning, and t/s denotes second per frame. For fair comparison, we initialize the backbone ResNet-50 pretrained on ImageNet \cite{imagenet2009A} instead of the MaskRCNN \cite{maskrcnn2017A} weights when only adopting DAVIS 2017 dataset for training.}
	\vspace{-1.0em}
	\begin{center}
		\setlength{\tabcolsep}{0mm}{
			\begin{tabular*}{\hsize}{@{}@{\extracolsep{\fill}}lcccccc@{}}
				\toprule[1.0pt]
				\multicolumn{7}{c}{\emph{Add YouTube-VOS for Trainin}g} \\
				\midrule
				Methods & S & FT & $\mathcal{J}$ & $\mathcal{F}$ & $\mathcal{J} \& \mathcal{F}$ & t/s \\
				\midrule
				AGSSVOS  \cite{agss2019A} & \checkmark & - & 64.9 & 69.9 & 67.4 & 0.10\\
				STMVOS  \cite{Seoung2019A} & \checkmark & - & 79.2 & 84.3 & 81.8 & 0.32\\
				EGMN  \cite{lu2020A} & \checkmark & - & 80.2 & 85.2 &  82.8 & 0.40\\
				KMNVOS  \cite{Seong2020A} & \checkmark & - & 80.0 & 85.6 & 82.8 & 0.24\\
				\midrule
				AGAME  \cite{agame2019A} & - & - & 67.2 & 72.7 & 70.0 & 0.14\\
				FEELVOS  \cite{feelvos2019A} & - & - & 69.1 & 74.0 & 71.5 & 0.51\\
				FRTM  \cite{frtm2020A} & - & - & 73.8 & 79.8 & 76.7 & 0.09\\
				LWL  \cite{Goutam2020A}  & - & - &  79.1 & 84.1 & 81.6 & 0.15\\
				CFBI  \cite{CFBI2020A} & - & - & 79.1 & 84.6 & 81.9 & 0.17\\
				CFBI$^{MS}$  \cite{CFBI2020A} & - & - & 80.5 & 86.0 & 83.3 & 9\\
				\textbf{JOINT (Ours)} & - & - & \textbf{80.8}& \textbf{86.2} & \textbf{83.5} & 0.25\\
				\midrule[1.0pt]
				\multicolumn{7}{c}{\emph{Only DAVIS 2017 for training}} \\
				\midrule
				Methods & S & FT & $\mathcal{J}$ & $\mathcal{F}$ & $\mathcal{J} \& \mathcal{F}$ & t/s \\
				\midrule
				OnAVOS \cite{onavos2017A} & - & \checkmark & 61.0 & 66.1 & 63.6 & 26\\
				AGSSVOS \cite{agss2019A} & \checkmark & - & 63.4 & 69.8 & 66.6 & 0.10\\
				RGMP \cite{rgmp2018A} & \checkmark & - & 64.8 & 68.6 & 66.7 & 0.28\\
				STMVOS \cite{Seoung2019A} & \checkmark & - & 69.2 & 74.0 & 71.6 & 0.32\\
				KMNVOS \cite{Seong2020A} & \checkmark & - & 74.2 & 77.8 & 76.0 & 0.24\\
				PReMVOS \cite{premvos2018A} & \checkmark & \checkmark & 73.9 & \textbf{81.7} & 77.8 & 37.6\\
				\midrule
				VideoMatch \cite{videomatch2018A} & - & - & 56.5 & 68.2 & 62.4 & 0.35\\
				FRTM \cite{frtm2020A} & - & - & 66.4 & 71.2 & 68.8 & 0.09\\
				LWL \cite{Goutam2020A} & - & - & 72.2 & 76.3 & 74.3 & 0.15\\
				CFBI \cite{CFBI2020A} & - & - & 72.1 & 77.7 & 74.9 & 0.17\\
				\textbf{JOINT (Ours)} & - & - & \textbf{76.0} & 81.2 & \textbf{78.6} & 0.25\\
				\bottomrule[1.0pt]
			\end{tabular*}
		}
	\end{center}
	\label{table:dv2017-val}
	\vspace{-2.5em}
\end{table}

\noindent\textbf{YouTube-VOS \cite{Xu2018YouTubeVOSAL}.}
YouTube-VOS is a large-scale benchmark for multi-object video segmentation which provides a much larger scale of training and test data than DAVIS. For the 2018 version, its validation set contains 474 videos, including 65 training (seen) categories and 26 unseen categories. And the 2019 version further augments the dataset with more video sequences, the number of videos in the validation set is increased to 507. The unseen object categories make the YouTube-VOS much suitable for evaluating the generalization capability of algorithms.

We evaluate the proposed approach on both versions of the YouTube-VOS benchmark. As shown in Table \ref{table:yt-val}, we compare our method with previously best-performing algorithms such as PReMVOS \cite{premvos2018A}, STMVOS \cite{Seoung2019A}, EGMN \cite{lu2020A}, KMNVOS \cite{Seong2020A}, CFBI \cite{CFBI2020A}, and LWL \cite{Goutam2020A}. We can observe that our approach achieves average scores of $83.1\%$ and $82.8\%$ on the two versions of the validation set respectively, which outperform other state-of-the-art methods by a considerable margin. Moreover, we discover that the generalization capability of our approach is significantly better than previous algorithms. For the training (seen) categories, the matching-based methods like KMNVOS \cite{Seong2020A} and CFBI \cite{CFBI2020A} perform well. However, when it comes to the performance on unseen categories, the previously mentioned methods plummet sharply, while our approach still maintains a relatively high level. Besides, our approach is superior to inductive learning-based methods such as LWL \cite{Goutam2020A} thanks to the spatio-temporal consistency exploration of our proposed lightweight transformer.

\noindent\textbf{DAVIS 2017 \cite{DAVIS2017}.}
DAVIS is a popular video object segmentation benchmark. The validation set of DAVIS 2017 contains 30 densely annotated videos, and it is more challenging compared with DAVIS 2016 \cite{DAVIS2016} since the multi-object setting is introduced. Follow \cite{Goutam2020A}, we split our approach into two versions depending on whether additional training data is employed or not and report their performance separately. We include the recently proposed LWL \cite{Goutam2020A}, CFBI \cite{CFBI2020A}, KMNVOS \cite{Seong2020A}, and EGMN \cite{lu2020A} for comparison. As shown in Table \ref{table:dv2017-val}, when additionally adopting YouTube-VOS for training, our method exhibits the best performance with an average ($\mathcal{J} \& \mathcal{F}$) score of $83.5\%$, outperforming all previous approaches in the literature. Compared with CFBI$^{MS}$ \cite{CFBI2020A} (enhanced version of CFBI), our approach is free of multi-scale and flip strategy during evaluation, thus runs 30+ times faster.


When only adopting DAVIS 2017 for training, we initialize the backbone ResNet-50 with ImageNet \cite{imagenet2009A} pretraining weights instead of the MaskRCNN \cite{maskrcnn2017A} weights for fair comparison. As we can see, in this setup, our approach still outperforms all previous methods with a $\mathcal{J} \& \mathcal{F}$ score of $78.6\%$.
Though the performance of PReMVOS \cite{premvos2018A} is close to ours, it relies on extensive online fine-tuning, so our approach runs about two orders of magnitude faster than it.
Note that methods like STMVOS \cite{Seoung2019A}, KMNVOS \cite{Seong2020A}, and PReMVOS \cite{premvos2018A} rely on additional synthetic data for pretraining. In contrast, our method is free of such necessity. 

\section{Conclusion}
In this work, we design a novel architecture for semi-supervised video object segmentation, which takes advantage of both transductive reasoning and online inductive learning. To bridge the gaps between the two diverse models and better exploit their complementarity, we adopt a two-head label encoder to generate disentangled mask encodings as the carrier of target information. Extensive experiments show that the proposed approach sets several state-of-the-art records on prevalent VOS benchmarks without the need of simulated training data.

\footnotesize {\flushleft \bf Acknowledgements}. This work was supported in part by the National Natural Science Foundation of China under Contract 61822208, 61836011, and 62021001, and in part by the Youth Innovation Promotion Association CAS under Grant 2018497. It was also supported by the GPU cluster built by MCC Lab of Information Science and Technology Institution, USTC.

{\small
\bibliographystyle{ieee_fullname}
\bibliography{egbib}
}

\end{document}